# A Methodology for Bi-Directional Knowledge-Based Assessment of Compliance to Continuous Application of Clinical Guidelines

**Authors:**

Avner Hatsek

Department of Software and Information Systems Engineering, Ben Gurion University, Beer Sheva, Israel

Declarations of interest: none

Yuval Shahar

Department of Software and Information Systems Engineering, Ben Gurion University, Beer Sheva, Israel

Declarations of interest: none

# Abstract

Clinicians often do not sufficiently adhere to evidence-based clinical guidelines in a manner sensitive to the context of each patient. It is important to detect such deviations, typically including redundant or missing actions, even when the detection is performed retrospectively, so as to inform both the attending clinician and policy makers. Furthermore, it would be beneficial to detect such deviations in a manner proportional to the level of the deviation, and not to simply use arbitrary cut-off values.

In this study, we introduce a new approach for automated guideline-based quality assessment of the care process, the *bidirectional knowledge-based assessment of compliance* (BiKBAC) method. Our BiKBAC methodology assesses the degree of compliance when applying clinical guidelines, with respect to multiple different aspects of the guideline (e.g., the guideline's process and outcome objectives). The assessment is performed through a highly detailed, automated quality-assessment retrospective analysis, which compares a formal representation of the guideline and of its process and outcome intentions (we use the Asbru language for that purpose) with the longitudinal electronic medical record of its continuous application over a significant time period, using both a top-down and a bottom-up approach, which we explain in detail. Partial matches of the data to the process and to the outcome objectives are resolved using fuzzy temporal logic.

We also introduce the *DiscovErr* system, which implements the BiKBAC approach, and present its detailed architecture. The *DiscovErr* system was evaluated in a separate study in the type 2 diabetes management domain, by comparing its performance to a panel of three clinicians, with highly encouraging results with respect to the completeness and correctness of its comments.

**Key Words**

**Highlights**

- A new top-down & bottom-up approach for automated guideline-based quality assessment
- Relies on comparing a formal representation of the guideline to the medical record
- Partial matches with the process and outcome intentions, using fuzzy temporal logic
- Implemented as the *DiscovErr* system, using Asbru for guideline representation
- Evaluated successfully in the type 2 diabetes management domain

# 1. Introduction

## 1.1. The Need for Automated Assessment of the Quality of Evidence-Based Medical Care

Clinicians and patients often do not sufficiently comply with evidence-based clinical guidelines. Examples include the only partial and highly variable compliance of physicians to type 2 diabetes management guidelines [Eder et al., 2019], and a compliance of only 41% to 49% to an American College of Obstetrics and Gynecology (ACOG) guideline for management of pregnant women who have Preeclampsia/Toxemia (PET), accompanied by an action redundancy rate of 68%. When guideline-based recommendations were suggested by a decision-support system, compliance rose to 93% and redundancy was reduced to 3% [Shalom et al., 2015].

It is important to detect such deviations, typically including redundant or missing actions, even when the detection is performed retrospectively, so as to inform both the attending clinician and policy makers. Furthermore, it would be beneficial to detect such deviations in a manner proportional to the level of the deviation, and not to simply use arbitrary cut-off values.

Multiple types of computational tools were developed in order to increase the compliance to the guidelines in medical settings and to assist clinicians to apply latest medical knowledge in real time. These tools include guideline search and visualization engines, frameworks for specification of guidelines in formal formats, and tools for application of guidelines knowledge for clinical decision support. A comprehensive review of most of these systems can be found in [Peleg et al. 2003; De Clerq et al. 2004; Peleg 2013].

In recent years, healthcare providers have invested increasing efforts in applying methods to assess the quality of the medical care that they provide for their patients. These efforts include the definition and publication of objective quantifiable quality measures that are being used as a guide for proper treatment and for evaluating the healthcare quality provided by clinical organizations and in private practice. Examples of such quality measures are the Clinical Quality Measures (CQMs) published by the Centers for Medicare & Medicaid Services (CMS), and the Indicators for Quality Improvement (IQIs) published by the NHS. Efforts are also being made for the development of automated systems for reporting the compliance to these measures, including the development of quality data models by organizations such as the National Quality Forum (NQF), for the linking of local databases to the standards of the publishers [Dykes et al. 2010], and for the development of tools to improve the quality of data in order to support automated analysis of these measures [Lanzola et al. 2014].

A growing number of medical organizations have established internal quality and risk assessment units that perform quality assessment of the medical treatment, typically on a random subset of the patient population. These risk assessment units usually examine the medical records manually, or by using relatively simple computational methods, and compare the medical records to a deterministic set of quality measures that are created specifically for that purpose. However, our objective in this study is to propose a methodology for automated assessment of the quality of care, and in particular, of evidence-based care.

## 1.2. Automation of Guideline-Based Critiquing and Quality Assessment

Several approaches were suggested over the past several decades, for the development of automated systems for guideline-based plan recognition, critiquing, and quality assessment. One of the first systems for medical critiquing was the HyperCritic system [van der Lei, Musen 1990], which examined electronic medical records and generated critiquing statements by applying a set of previously defined critiquing-tasks. The critiquing process that was implemented in the HyperCritic system used formally represented knowledge that supported the application of a set of critiquing tasks. The knowledge model comprised two distinct types of knowledge, *critiquing knowledge* and *medical knowledge*. The critiquing knowledge was used by the system to determine when to apply the critiquing and how the critiquing should be applied, whereas the medical knowledge described the medical facts that are used during the application of the critiquing tasks. The HyperCritic system was implemented in the hypertension domain, and was later extended via a new implementation in the domain of asthma and chronic obstructive pulmonary disease (COPD), called AsthmaCritic [Kuilboer et al. 2003].

Note that such retrospective critiquing systems provide useful commentary on the manner in which the care provider has implemented a particular care plan, using solely the electronic medical record's *data*. This type of retrospective-critiquing systems complements the real-time critiquing systems, which view the care provider's *plan*, but are less intrusive than prescriptive guideline-based decision-support systems (which suggest evidence-based recommendations), since they produce comments only when significant problems are detected in the care-provider's plan [Miller 1986].

Meanwhile, Reid Simmons had been developing an intriguing approach to the interpretation of a current state of affairs, as a possible transformation, or revision, of an original plan using a Generate, Test and Debug (GTD) paradigm [Simmons 1988]. Inspired by Simmons' approach, Shahar and Musen [1995] presented several fundamental aspects required to be addressed before such systems can be adopted, such as the need for recognition of the plans of the human care provider and their objectives, the need for applying mechanisms of plan revision to better understand how the original plan might have been transformed, and the importance of a sophisticated temporal reasoning engine, to enable the analysis of the time-based data that are collected on each patient. An additional system for medical critiquing was presented by Gertner [1997], who developed the Trauma-TIQ system that extended the Trauma-AID [Clarke et al. 1993] system. The Trauma-TIQ system was aimed to assist in decision support for trauma management by critiquing the clinician's actions only when significant gaps from the guideline were detected by system.

Several years later, Sips et al. [2006] presented an algorithm for an intention-based matching process and its evaluation in cases of hyperbilirubinemia. Another system, IGR, was presented by Boldo [2007] for plan-recognition of clinical guideline plans, by analyzing the data in the electronic medical records. To perform the plan recognition, the system used an abstract format to represent the clinical guidelines, called guideline characteristic vectors. The plan recognition method of the IGR system included the use of fuzzy logic techniques to support partial matching between patient data and the guidelines. Panzarasa et al. [2007] presented the RoMA module for analysis of motivations for non-compliance of clinicians to guideline recommendations presented to them by a care-flow system. Groot et al. [2008] proposed an additional computational method to perform the critiquing, by representing the actual treatment actions and the clinical guidelines using temporal logic and a state transition system, and employing model checking to investigate whether the actual treatment is consistent with the guidelines. Several researchers had developed platforms for detecting whether a set of quality indicators is satisfied, given a patient's record. For example, Hazelhurst et al. [2012] have developed and evaluated, in two different health systems, a system that assesses the quality of outpatient care of asthma, using 22 quality measures.

Advani et al. [2002] have outlined an approach for quality assessment by scoring adherence to the guideline's intentions, and presented the Quality Indicator Language (QUIL), which was based on the

Asbru language [Shahar et al. 1998] and was specifically designed to represent knowledge for the quality assessment task. In the current study, we have decided to exploit the Asbru language due to its highly explicit representation of time-oriented properties of clinical guidelines, and its explicit specification of the guideline's objectives at multiple levels, which support our main task.

**The Asbru Guideline-Representation Language**

*Asbru*, which we had exploited also in our current study, is an expressive guideline-representation language that enables guideline designers to represent procedural sequential, concurrent, conditional, and repeating actions as a hierarchy of execution plans. To control the application of the plans, Asbru uses multiple types of entry conditions (specifically, compulsory *filter conditions* such as age or gender that must be true on entry, and achievable *setup conditions* such as having access to the patient's glucose-tolerance test, which can be satisfied using some action or plan), as well as *abort*, *suspend*, *restart*, and *complete* conditions, represented as temporal-abstraction patterns (see below). Furthermore, to support quality assessment, Asbru includes a meta-level of explicit intermediate and overall *process* (e.g., administration of a beta blocker twice a day) and *outcome* (e.g., reduction of diastolic blood pressure below 85 mmHg) *intentions*, also represented as temporal-abstraction patterns, whose objective is to *maintain* (if they already exist), *achieve* (if they are currently false), or *avoid* (assuming they are false and should stay so) some temporally extendable goal [Shahar et al., 1998; Miksch et al., 1997]. The temporal semantics of Asbru have also been shown to be consistent and clear [Schmitt et al., 2006].

Asbru has been used in multiple clinical decision-support projects, such as in the MobiGuide EU project, in which it was used to represent a gestational-diabetes and gestational hypertension guideline to manage patients in Spain, and an atrial-fibrillation guideline to manage patients in Italy [Peleg et al., 2017a, 2017b]. The BiKBAC compliance-assessment method assumes that the guideline is formally represented in Asbru or in some other language equivalent to it (i.e., which includes a semantic equivalent to all of Asbru's actions, conditions, and intentions).

**The *Knowledge-Based Temporal-Abstraction* (KBTA) Method**

All conditions and actions and intentions in Asbru are based on domain knowledge that is represented as temporal-abstraction patterns. *Temporal abstractions* are interval-based, abstract concepts (e.g., "moderate anemia for three weeks") derived from the input raw, time-stamped clinical data (e.g., a series of point-based Hemoglobin-value measurements).

In our study, we have derived all temporal abstractions from the raw time-stamped data using domain knowledge that is formally represented using the ontology underlying the *knowledge-based temporal-abstraction* (KBTA) method [Shahar, 1997; Shahar and Musen, 1995], and used the KBTA method to compute these temporal abstractions, such as to decide whether a particular entry condition holds, or whether an outcome intention was achieved.

The KBTA ontology defines measurable *raw concepts* (e.g., blood-glucose value); interval-based *abstract concepts*, which include abstractions of type *State* (e.g., High and Low value), *Gradient* (e.g., Increasing or Decreasing or Stable blood pressure), *Rate* (e.g., changing quickly), *Trend*; *Events*, which include external, usually volitional actions such as administration of a medication or a surgical procedure; *Patterns* (e.g., proper administration of a certain medication), which are interval-based combinations of events, raw and abstract concepts, and possibly other patterns, which need to satisfy various value and time constraints for each interval and among the intervals; and *interpretation contexts* (e.g., the period during which short-acting insulin affects blood glucose), which are *induced* by all other ontological entities (e.g., for 30 minutes and up to eight hours following the administration of a regular insulin injection), and change, in a context-sensitive manner, the specific knowledge (e.g., the *functional knowledge* that determines the definition of a High blood-glucose state that is applied in that context, such as before or after a meal). Several types of temporal-abstraction knowledge exist - structural, functional, logical, and probabilistic [Shahar, 1997].

In the Methods Section 2.2, we describe a fuzzy extension of the KBTA method, the fuzzy temporal reasoner, which assigns a membership value between 0 and 1 to each abstraction with respect to fulfilling a certain condition or intention.

**Main Contributions of the Current Study**

There have been several approaches to the task of providing a critique regarding evidence-based therapy. However, none of the current approaches exploits a full formal representation of the clinical guideline, nor do these approaches apply quality measures that exploit full-blown temporal patterns that represent a correct guideline-based process being carried out as planned, in a manner sensitive to the longitudinal, evolving contexts of each patient, including all of the guideline's intermediate and overall process and outcome objectives, and catering also for partial (temporal) pattern matching of these temporally oriented objectives. Partial temporal-pattern matching is important, since it would be highly beneficial to detect deviations in compliance to a given guideline in a manner proportional to the level of the deviation, and not to simply use arbitrary cut-off values.

The methodology that we are introducing in the current study, the *Bidirectional Knowledge-Based Assessment of Compliance* (BiKBAC) method, fulfils these requirements. The new methodology exploits a formal representation of the evidence-based guideline, and uses both a *top-down* (driven by the guideline's process and outcome intentions, represented formally as Asbru temporally extended intentions) and a *bottom-up* (driven by the patient's multivariate longitudinal data) approach. The need for avoiding overly crisp cut-off values is addressed in the BiKBAC methodology through the performance of partial matches of the data to the process and to the outcome objectives of the guideline, which are resolved using fuzzy temporal logic.

In this paper, we describe in detail the BiKBAC methodology. We also introduce the *DiscovErr* system, a full implementation of the BiKBAC methodology. The BiKBAC methodology and its implementation as the *DiscovErr* system are the main contributions of the current paper. The *DiscovErr* system has been evaluated rigorously in a study, which we briefly summarize, that is outside of the scope of the current paper. The system has been shown to have a quality-assessment performance level, with respect to completeness and correctness, equivalent to that of medical experts in the particular clinical domain (management of type 2 diabetes patients) that was used in the evaluation. Thus, the BIKBAC approach has also been shown to be highly valid.

## 2. Methods
### 2.1. The Bidirectional Knowledge-Based Compliance-Analysis Methodology

We shall start by introducing our computational method, the *Bidirectional Knowledge-Based Assessment of Compliance* (BiKBAC) method, and the architecture of the *DiscovErr* system implementing it.

The BiKBAC methodology assumes that the guideline is represented in a formal fashion at two levels. These levels include (1) its overall process (recommended actions) and its transition conditions (e.g., *entry* condition, *stop* condition) and (2) its *process* and *outcome* intentions. The representation need not be identical, but should be expressively equivalent to that of the Asbru guideline-representation language, as explained in Section 1.2. For example, the subtypes of the entry condition must include some version of the Asbru *Filter* (i.e., compulsory) and *Setup* (i.e., a state to be achieved) entry conditions, while the subtypes of the stop condition must include some version of the Asbru *Complete*, *Abort*, and *Suspend* conditions, with similar semantics to those of the Asbru language. Both of these conditions, as well as the process and outcome intentions, are represented formally using the KBTA ontology and computed using the KBTA method, as explained in Section 1.2.

The goal of this compliance analysis methodology is to evaluate the data of each single patient with respect to its compliance with the process and outcome objectives of the full set of operative formal clinical guidelines that are represented in the system's library, and in particular, those guidelines that are relevant to the patient's state at each point in time. When invoked, the algorithm accepts the complete set of patient's data as an argument; the data are represented using an in-memory object that holds both the demographic and the temporal data of the patient.

The algorithm that is at the core of the BiKBAC method combines two computational approaches: *top-down* and *bottom-up*. The algorithm consists of several sequential steps, in which each step can directly analyze the raw patient data or use the outputs calculated in the previous steps of the algorithm. The outputs of the steps are stored in a central data structure called a *TimeLine*, which supports fast storage and retrieval of the time-stamped data items. The main steps of the algorithm and their respective order are presented in Figure 1 and are described in the following sections.

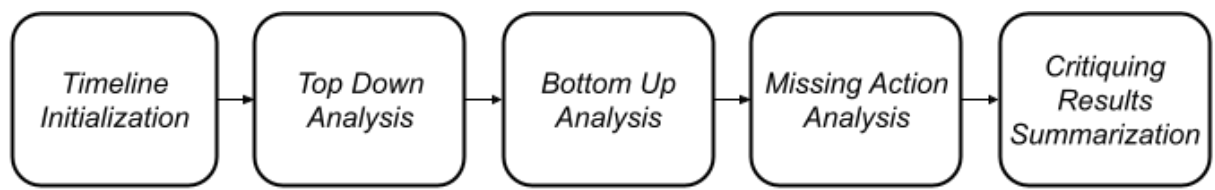

**Figure 1.** The key steps in the Bidirectional Knowledge-Based Assessment of Compliance (BIKBAC) method

The TimeLine data structure is used to store and retrieve time-stamped data items, which are called *TimePoints*, which represent multiple events in the patient's history. Each TimePoint includes an annotation regarding its type, a timestamp, and references to additional relevant information according to its type. The TimePoints are divided into two main types: Data-Item points represent the raw data items that exist in the original medical record, while Computed-Explanation points represent new higher-level information that is created during the compliance analysis. The description of the core algorithm of the BiKBAC method in the following sections clarifies the context in which the various types of computed-explanation points are added to the TimeLine.

**Top-Down Analysis**

The Top-Down analysis is a computational process that analyzes the time-stamped patient data from the perspective of the set of potentially relevant guidelines, i.e., a *knowledge-driven* process. In general, it is performed for the detection, within each time-stamped patient data series, of the high-level interval-based temporal abstractions that are related to the entry conditions, stop conditions, and/or outcome intentions of any operative guideline that is available in the knowledge library.

Figure 2 presents a flowchart illustrating the steps of top-down analysis; a description of the process is provided in the following paragraphs.

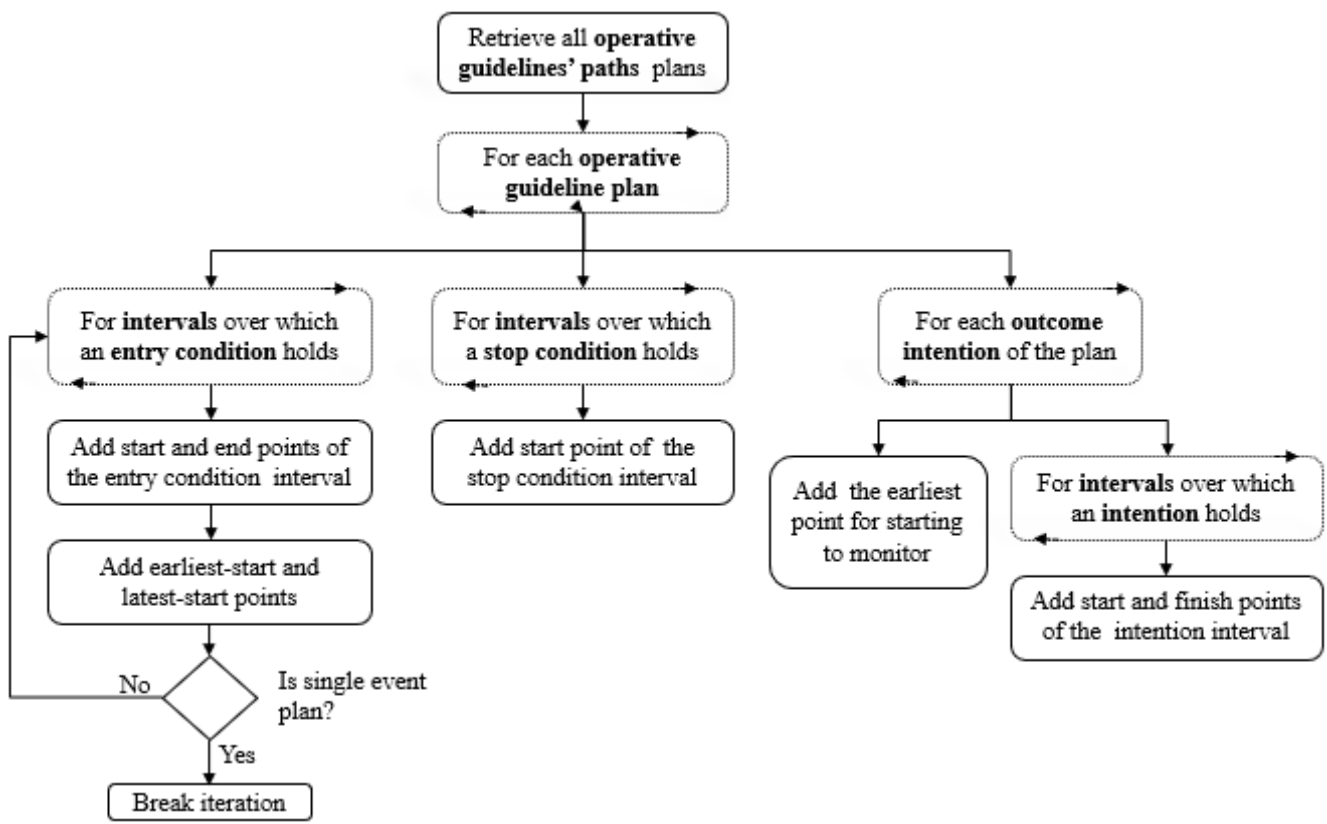

**Figure 2.** The top-down Analysis step of the BiKBAC method's compliance analysis algorithm. The dashed rectangles represent iterative steps performed on a specific collection. The top-down analysis is performed on each operative guideline plan in the library.

When starting the top-down analysis, the system retrieves the complete set of operative guideline plans, and then breaks composite guidelines into a virtual set of guideline-plans, each describing a single clinical path of the original composite guideline. The virtual plans are assigned the guideline-conditions (i.e., Asbru's conditions: *Filter, Setup, Complete, Abort, Suspend*, and *Restart*) of the original "parent" guideline. We call this process "*Condition Propagation*", and it involves the following logic: When propagating an entry condition of a guideline (i.e., *Filter, Setup,* and *Restart*) from a parent plan to one of its sub-plans, the condition of the parent plan is added as a conjunction (i.e., AND relation) to the existing entry condition of the sub-plan, for example "Diabetes AND Pregnancy". That is, *all* of the entry conditions must hold to enter the sub-plan, including those of its parent plan. When propagating a stop condition of a guideline (i.e., *Complete, Abort,* or *Suspend*) from a parent plan to one of its sub-plans, the condition of the parent plan is added as a disjunction

(i.e., OR relation) to the existing stop condition of the sub-plan, for example "Patient Hospitalization OR Severe Pain". That is, it is sufficient that any of the stop conditions will hold, including one of those propagated from the parent plan, to stop the sub-plan.

For each operative guideline path, the declarative concepts of the entry conditions, stop conditions, and outcome-intentions are extracted and sent to the *Fuzzy Temporal Reasoner* (which we describe in section 2.2), which evaluates them on the patient data. For each of these declarative concepts, the Fuzzy Temporal Reasoner returns a set of zero or more temporal abstractions, in the form of time stamped intervals assigned with a membership score (from 0 to 1). The system filters the returned intervals and accepts only those with a membership score that passes certain configurable thresholds. The remaining intervals are then used for adding data points to the TimeLine according to the logic described in the flowchart in Figure 2.

Thus, for example, in the case of the type 2 diabetes management guideline, an entry condition (in this case, a compulsory Filter condition) of HbA1C level greater than 6.5%, would be used to label time intervals during which this condition held, indicating also to what (fuzzy) degree the condition held (on a scale of 0..1).

**Bottom-Up Analysis**

The Bottom-Up analysis is a computational process that analyzes the guideline compliance from the perspective of the patient's data, i.e., a *data-driven* process. In general, the bottom-up analysis consists of a process that examines each data item in the patient's medical record, and provides it with a set of possible knowledge-based (i.e., clinical guideline-based) explanations.

In the bottom-up analysis, the system scans each item in the patient data, and tries to provide as many computed explanations for each data item as possible. The computed explanations are structured semantic comments that evaluate the data item in the context of any guideline-plan that is related to this type of items, i.e., the item is included in the formal definition of a certain part of the guideline. This, of course, is done according to what is known to the system, i.e., the full set of formal operative guidelines.

For example, consider a hemoglobin A1c measurement that is potentially related to a diabetes guideline in two ways: It can be a pre-diabetic (screening) test that is taken to decide if the patient has diabetes (included in the guideline's filter-condition), or it can be an ongoing periodic test step (included in the guideline's plan-body), to monitor a patient who was already diagnosed. The data item is examined in the context of these two different guideline knowledge roles, and a possible computed explanation is provided for each. Each of these computed explanations that are added for a specific data point, is represented in a data structure that holds additional details. For example, computed explanations in the context of guideline step points, are assigned a quality score for the assessment of the timing of the action, with a corresponding description of the temporal relation, which can be assigned step-too-early, step-on-time, or step-is-late. Note that each data point that represents a clinical parameter in the patient's TimeLine, is assigned with a collection of one or more possible computed explanations. In a later step of the compliance assessment, the computed explanations are summarized to select the most reasonable one. Figure 3 presents a hierarchical flow chart illustrating the main flow of the bottom-up analysis.

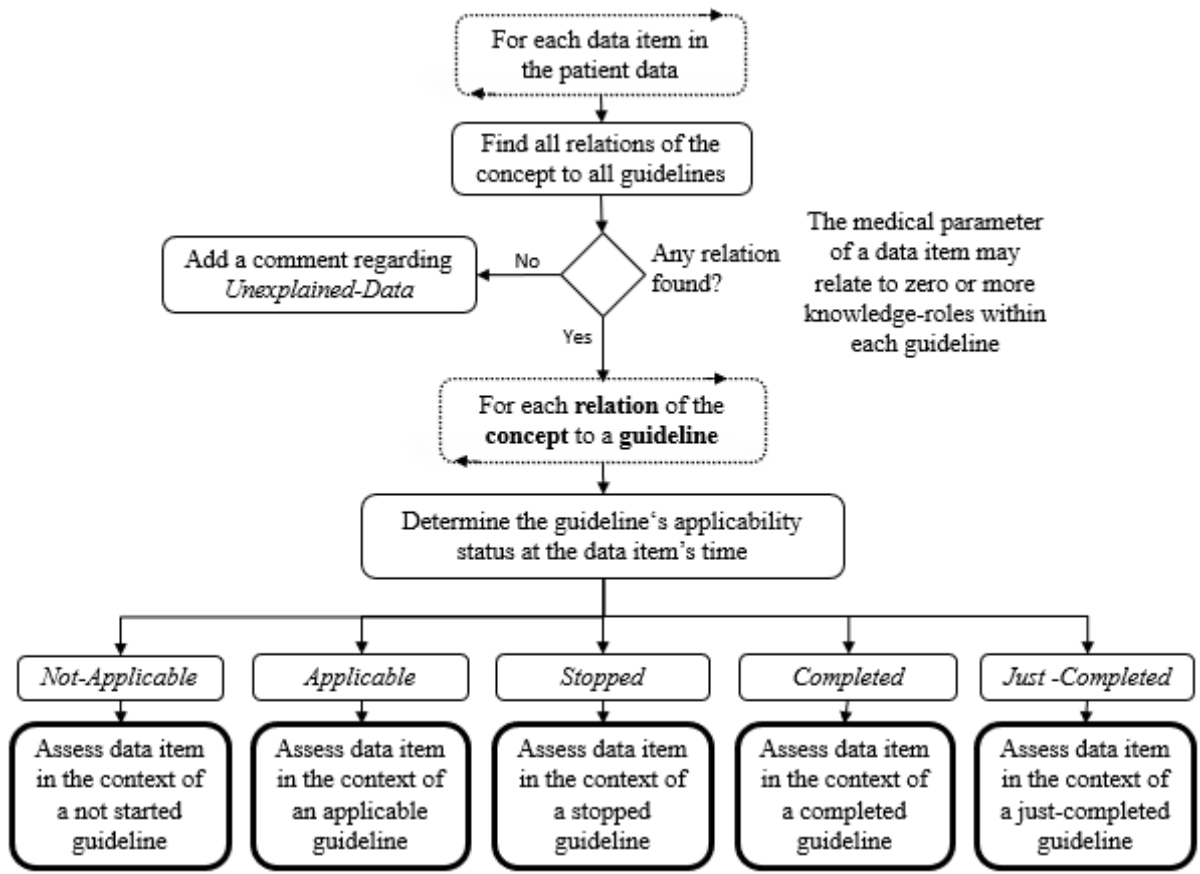

**Figure 3.** The bottom-up Analysis step of the BiKBAC method's compliance analysis algorithm.

In the next step, for data items that were found with relations to the knowledge roles (e.g., conditions) of potentially relevant operative guidelines, each of the relations is examined to determine the Applicability-Status of all potentially applicable guidelines. The applicability status can be generally described as an answer to the question "Was the guideline's plan applicable in the time referred to by the data item?", and is determined by scanning the previous events recorded in the TimeLine during the Top-Down analysis phase. These events include the guideline's (earliest and latest) start and stop points, which were added in the Top-Down analysis (see Figure 2), and the computed explanations of all of the clinical steps that occurred prior to the current examined time (recall that the data items are scanned chronologically). The applicability-status is used to determine the context in which to further analyze the data item, in which additional logic is applied to determine the type of explanation to add. For example, in the context of a Stopped-Guideline, as determined during the Top-Down analysis, a plan body item of this guideline, found in the Bottom-Up analysis, would be assigned the explanation Stopped-Guideline-Step, whereas in the context of a previously determined Completed-Guideline, the same plan body item would be explained after the Bottom-Up analysis as Redundant-Step-Repeated.

Thus, for example, while critiquing the management over time of a diabetes type 2 patient, (1) a HbA1C measurement result might be associated with the Filter Condition and the Outcome Intention of the type 2 diabetes management guideline (assuming that guideline exists in the guideline library); (2) an LDL cholesterol measurement result might be associated with the Filter Condition and the Outcome Intention of a Hyperlipidemia management guideline, if that guideline also happens to be in the guideline library; and (3) a Lithium medication (possibly given to control the patient's Bipolar disorder) would be labeled as "unexplained data" if the Bipolar Disorder guideline, or any other clinical guideline that involves administration of Lithium, does not exist in the guideline library.

**Missing Actions Analysis**

The Missing-Actions-Analysis, illustrated in Figure 4, is an important step of the compliance analysis, due to the fact that missing an action is one of the more common problems with respect to compliance to clinical guidelines. In this step, the system scans the TimeLine again, this time to detect missing actions. There is a need for an additional scan of (i.e., a third pass over) the TimeLine, due to the fact that in the previous steps of the compliance analysis the system could not detect the missing actions; in the Top-Down analysis, the system examines the guidelines' conditions and outcome-intentions but does not directly consider the clinical actions (which are represented in the plan-body of the formal guidelines); in the Bottom-Up analysis, the process scans the existing data items in the patient's record, thus, in this manner, it cannot detect actions that are missing. While scanning the TimeLine chronologically, when detecting a Plan-Latest-Start of a guideline, a Missing-Action comment is generated for each missing clinical action of that guideline.

In the specific and important case of assessing a missing Drug-Increase clinical steps item, the system is trying to explain the missing actions by considering two common scenarios in which the medication-increase action should be canceled even if it is required according to the patient's data. The first scenario is reaching a maximal dose of a medication, a situation in which a further dose increase is not recommended. The second scenario is cases in which the compliance to the medication was low to begin with, at the time of the assessment, meaning that the patient did not take the medication; in such a situation, it is more reasonable to improve the core compliance to the taking of the medication, rather than to increase the dose of the medication, and thus, that should be the output of the quality-assessment process.

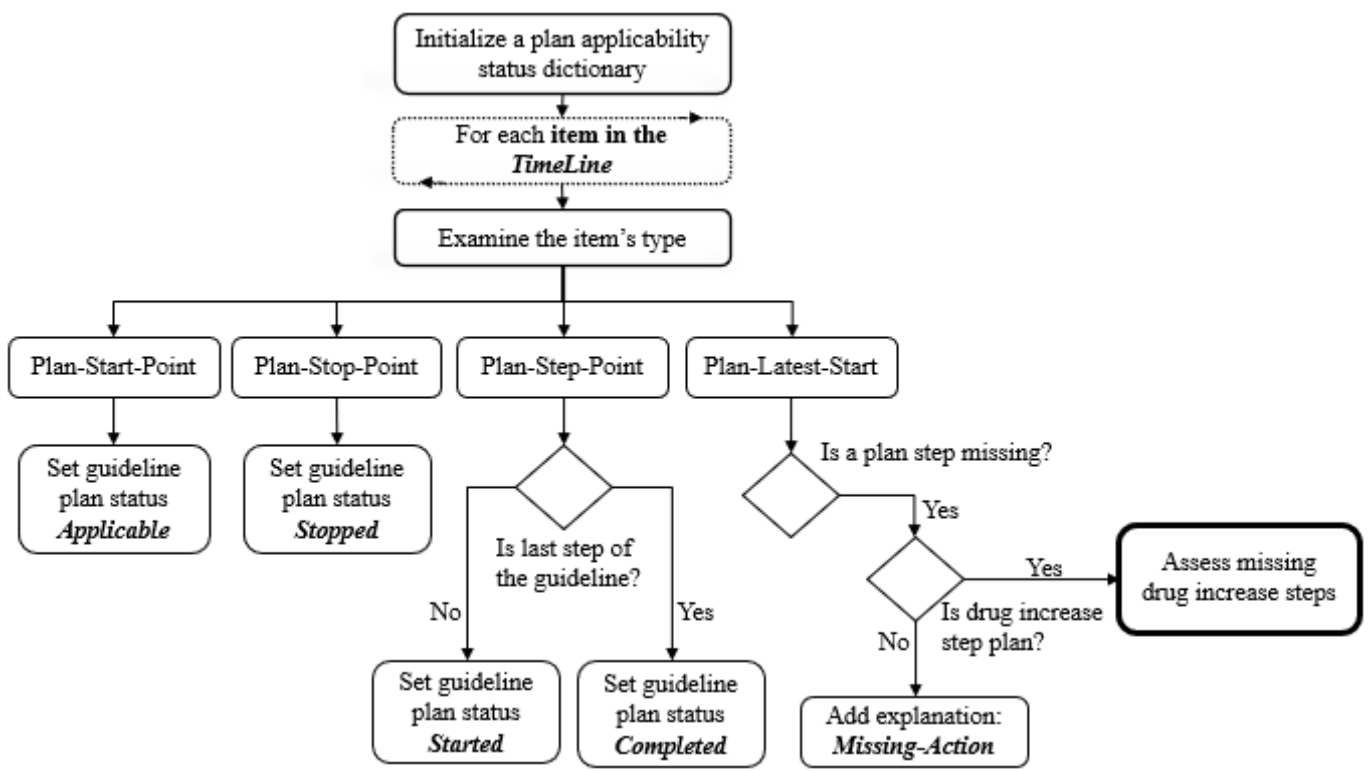

**Figure 4.** The Missing-Actions Analysis of the BiKBAC compliance-analysis methodology.

**Critiquing Results Summarization**

During the application of the previous steps of the compliance analysis algorithm, the system is generating a large amount of computed explanations, which are represented as TimePoints objects that are stored in the TimeLine. Some of these computed explanations represent useful comments regarding compliance with one or more relevant clinical guidelines (e.g. a late drug administration action), whereas other computed explanations are less useful for potential users and relevant mostly to support the analysis process itself (e.g. an internal notion of the system regarding a detected drug-increase action). In this phase, the system extracts the useful computed explanations from the TimeLine, and filters out the less useful ones. The useful computed explanations represent the system's Comments regarding compliance to the guidelines. The Comments are the BiKBAC method's main output.

Intention Related Comments are extracted from the outputs of the *top-down* analysis, in which each outcome-intention of a guideline-plan was evaluated using the Fuzzy-Reasoner. The outcome intentions-related explanations are represented as scored temporal-intervals, and for each outcome-intention the following sets of temporal-intervals are extracted: a set of temporal-intervals with assessment regarding the achievement of the intention (e.g., the hemoglobin A1c goals were almost on target in a certain period of time, therefore, the fuzzy membership score might be, say, 0.85); a set of temporal-intervals in which the intentions should have been monitored (e.g., for a patient who was diagnosed with diabetes in January, the outcome-intentions should have been monitored from April); and a set of temporal intervals in which the system detects insufficient data to determine the achievement of the outcome intention, i.e., intervals in which an intention was not monitored.

Data Item Comments are extracted from the outputs of the *bottom-up* analysis, in which, as explained, each data item in the patient record is evaluated and assigned zero or more computed explanations. In general, it is reasonable to provide multiple explanations when analyzing data in a retrospective manner, however, for practical reasons, there is a need to organize these multiple computed-explanations in a manner that emphasizes the most reasonable computed explanations provided for each data item. The computed-explanations of each data item are sorted using a score that represents the "reasonableness" (i.e., likelihood of being relevant) of the computed explanations. The computed explanation with the highest "reasonableness" score is selected as the compliance comment and assigned to the examined data item. The "reasonableness" score is the mean of the following scores given to each explanation: (1) a score that represents the level of applicability of the guideline at the valid time of the data item; (2) a score that describes the strength of the relation between the data item to the knowledge role of the relevant clinical guideline (e.g., the score of a data item unique to a particular guideline is higher, for that particular guideline, than the score of a data item that might be equally relevant to three different guidelines, with respect to each of the three guidelines); (3) a score for the timing of the clinical step that is represented by the data item, available only for clinical steps that were expected according to an applicable guideline.

The full set of data item comment types include the following: step-not-supported (by an applicable known guideline), stopped-plan-step, redundant-step-repeated, duplicate-step, wrong-path-selection (another guideline path is more suitable), step-too-early, step-on-time, step-too-late.

Missing Actions Comments are extracted from the output of the Missing-Actions-Analysis, in which the system scans the TimeLine and adds computed explanations regarding any missing action of the operative guideline. Extracting these computed explanations is important, since, as noted above, missing actions are a common problem of guideline compliance.

## 2.2. The Fuzzy Temporal Reasoner

Recall that the Top-Down analysis exploits fuzzy temporal reasoning to assign values to the degree to which certain conditions or intentions hold over time. The *Fuzzy Temporal Reasoner* is a KBTA-based engine that is used to extract high level interval-based interpretations from the raw, time-stamped temporal data, and assign fuzzy membership scores to the resultant abstractions. It uses formal declarative knowledge about medical concepts, which are represented according to the KBTA ontology. In addition, the Fuzzy Temporal Reasoner applies techniques of fuzzy logic [Zadeh 1965, 1968] that take part in the reasoning process, enabling it to provide a (fuzzy) *membership score* for each abstraction it generates. The use of fuzzy logic techniques distinguishes the Fuzzy Temporal Reasoner from other existing KBTA engines that are able to extract from raw data only deterministic abstractions based on classical First Order Logic.

In addition to applying temporal logic for the evaluation of the various types of KBTA abstract concepts (e.g., State, Gradient, Rate), the Fuzzy Temporal Reasoner uses techniques of fuzzy logic both in the evaluation of logical relations (e.g., $x < y$), and in the evaluation of logical operators (i.e., AND, OR operators) that are part of compound logical expressions.

The Fuzzy Temporal Reasoner is an internal component of the Analysis Framework and is used by the guideline Compliance Analysis Engine to evaluate the state of the patient with regard to the conditions specified in the guidelines. For example, when evaluating the compliance to a guideline regarding a medication that should be stopped in the case of reduced kidney function, the Compliance Analysis Engine uses the Fuzzy Temporal Reasoner to retrieve the time intervals during which the reduced kidney function state was equal to True according to the patient's data (i.e., the State abstraction "reduced-kidney-function-state" = 'True'). The results of the Fuzzy Temporal Reasoner include a set of time-intervals, each assigned with a membership score represented as a continuous number between 0 and 1. In time periods during which the raw data of the patient completely satisfy the constraints specified in definition of the "reduced-kidney-function-state" concept, the membership score is assigned the value of 1; during periods when these constraints are fully contradicted, the membership score is assigned the value of 0; during periods in which these constraints are partially satisfied, due to raw data that are close to satisfying the relation, the membership score is assigned a rational number between 0 and 1.

To better understand the motivation behind the use of fuzzy logic in the compliance analysis process, consider the following example of a CQM measure in the domain of the current study: "Low Density Lipoprotein (LDL-C) Control in Diabetes Mellitus", defined as "Percentage of patients aged 18 through 75 years with diabetes mellitus who had most recent LDL-C level in control (less than 100 mg/dL)". A simplistic algorithm using this rigid cut-off value would assign a care provider who manages a group of 50 patients, whose LDL-C values at the point in which they were examined were all just slightly higher than 100 mg/dL, say 106, a quality measure of zero. Such an extreme assignment is not likely to be well received by the clinical community, and would decrease trust and acceptance of quality assessment results. However, due to the use of the fuzzy temporal logic mechanism to assess quality measures, the *DiscovErr* system would assess the quality of care of that group as quite high, say, by assigning them a membership score of, say, 0.88 or even 0.93, although the compliance would not be perfect. However, an LDL-C value of, say, 130 or higher would be assigned a membership score of 0, while intermediate values would be assigned some membership score between 0 and 1, using a linear or another membership function to interpolate between the two extreme values.

The reasoning process that is applied by the Fuzzy Temporal Reasoner involves a multistep process. In the following paragraphs, we describe each step of the reasoning process, and demonstrate it using an example for the evaluation of a Hypertension (High Blood Pressure) concept. For this example, we define Hypertension as a *State* concept, abstracted from two *Primitive* parameters, systolic blood pressure (SBP) and diastolic blood pressure (DBP), each is a numeric parameter defined with a local persistency of one hour (i.e., the measurement is good for one hour). The abstraction includes the logic OR operator applied on two constraints on the values of the *Primitive* parameters. The definition of the Hypertension concept:

Hypertension = {SBP > 140 mmHg OR DBP > 90 mmHg}

Examples of the raw measurements of the patient's systolic and diastolic blood pressures are illustrated in Figure 5.

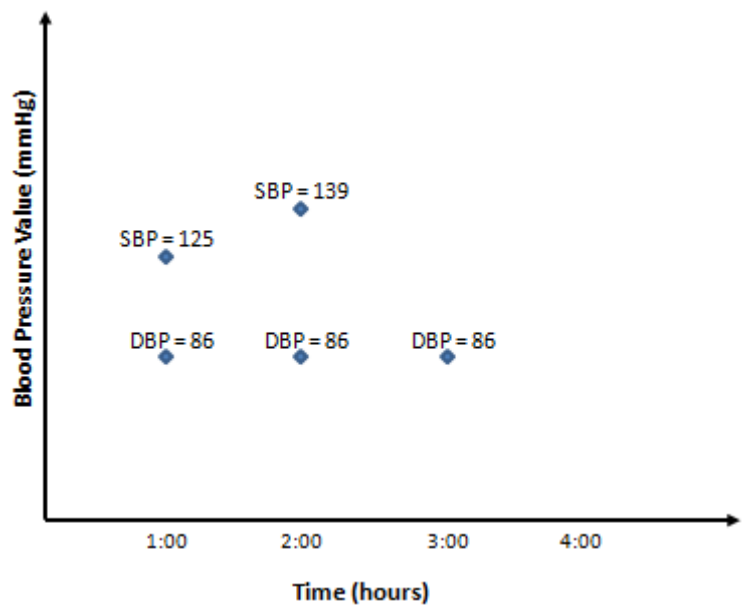

**Figure 5.** Blood Pressure measurements used for the demonstration of the Fuzzy Temporal Reasoner.

**Extrapolation of Temporal Intervals**

The first step of the reasoning process includes extrapolation of temporal intervals using the *Temporal-Persistence* knowledge that was specified for the raw parameters. In addition, each pair of intervals of the same parameter that share the same value, and share a mutual time period (i.e., overlapping intervals), are merged into a single interval of this parameter that is assigned the same value.

In the example (see Figure 6), each blood pressure measurement is now extrapolated to create an interval of one hour, the three DBP measurements are extrapolated to three consequent intervals with the same value = 86. In the merging step, these three intervals are merged into a single longer interval with the same value.

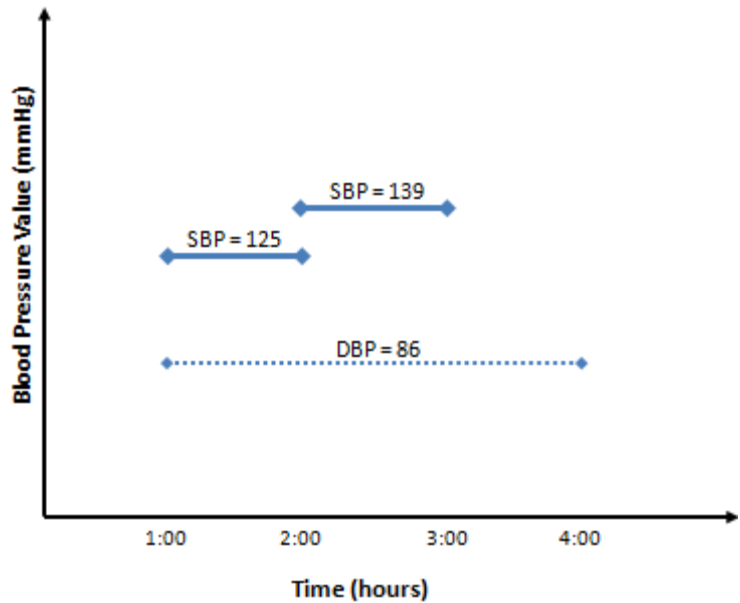

**Figure 6.** Extrapolation of the time-stamped measurements by the Fuzzy Temporal Reasoner.

**Temporal Partitioning**

The second step of the temporal reasoning process involves partitioning of the temporal data. The idea behind this operation is to create a segmented (partitioned) temporal representation, with the minimal set of partitions, in which each relevant parameter has zero or a single value. This is done to support the next steps of the reasoning process, in which the evaluation logic is applied on each of these partitions.

In the example (see Figure 7), the data is partitioned into five partitions. In the first partition, none of the parameters is provided with a value; in the second partition, the DBP value is 86, and the SBP value is 125; in the third partition, the DBP value is 86 and the SBP value is 139; in the fourth partition the DBP value is still 86 and the SBP has no value; and in the last partition, none of the parameters is provided with a value.

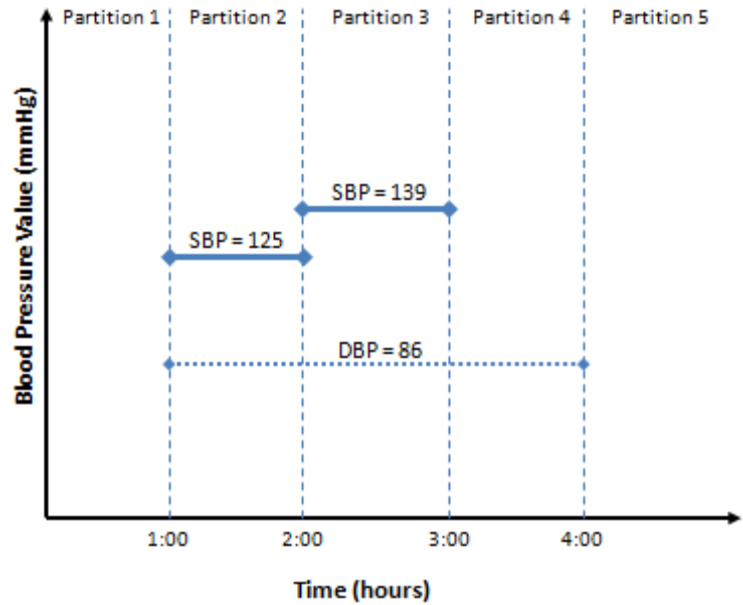

**Figure 7.** Temporal partitioning by the Fuzzy Temporal Reasoner.

**Fuzzy Evaluation of the Logical Relations**

The third step of the reasoning process involves the evaluation of the logical relations. The evaluation is done for each parameter value in each of the partitions.

In classic logic, the evaluation of logical relation results with a *true* or a *false* value, referred as the *truth-value* of a relation. In fuzzy logic, the result of a relation evaluation is called the membership score, and is represented as a continuous (rational) number, usually between 0 and 1. For that, we extended the representation of logical relations of the KBTA schema, with an attribute called a *deviation-interval*. This attribute is used in the relation evaluation to enable the assignment of a membership score in cases where the constraint is not fully satisfied. In such cases, a special *fuzzification-function* is applied for the reasoning process.

The *fuzzification-function* receives the following arguments: the *current-value* of the parameter, the *threshold*, the *deviation-interval,* and the *relation-operator* specified in the constraint definition. The *deviation-interval* represents the maximal deviation from the threshold that can be evaluated with a membership score higher than zero; thus, in cases where the absolute distance between the *current-value* and the *threshold* is greater than the *deviation-interval*, the membership score is evaluated as zero. In all other cases, the membership score is calculated according the following formula:

Figure 8 illustrates the application of the *fuzzification-function* for the evaluation of the constraint SBP > 140 mmHg, with a *deviation-interval* of 10 mmHg.

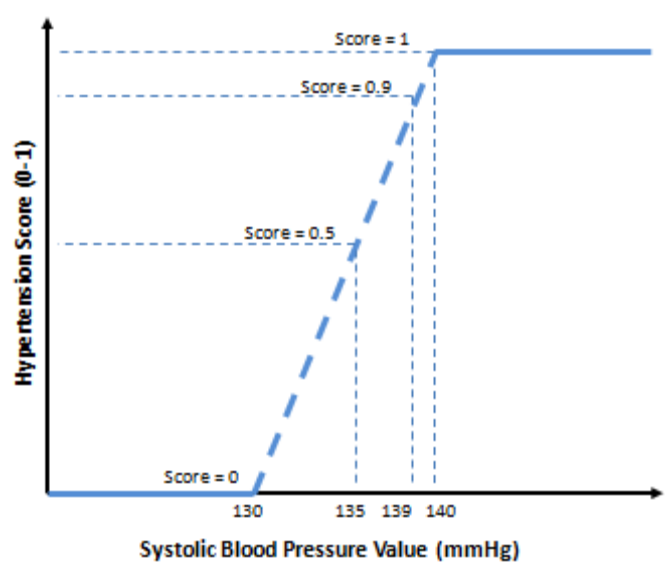

**Figure 8.** Illustration of the *fuzzification-function.* Evaluation of the constraint SBP>140 mmHg, with a deviation-interval of 10 mmHg. On a measurement of SBP=139, the membership score is evaluated as 0.9; on a measurement of SBP=135, the membership score is evaluated as 0.5; on any measurement of SBP≤130, the membership score is evaluated as 0; on any measurement of SBP≥140, the membership score is evaluated as 1.

Figure 9 illustrates the complete logical constraint evaluation on each parameter value in each partition that was generated in the previous step of the example. The interval with SBP=125 was evaluated with a membership score = 0, the interval with SBP=139 was evaluated with a membership score = 0.9, and each interval with DBP=86 was evaluated with a membership score = 0.6.

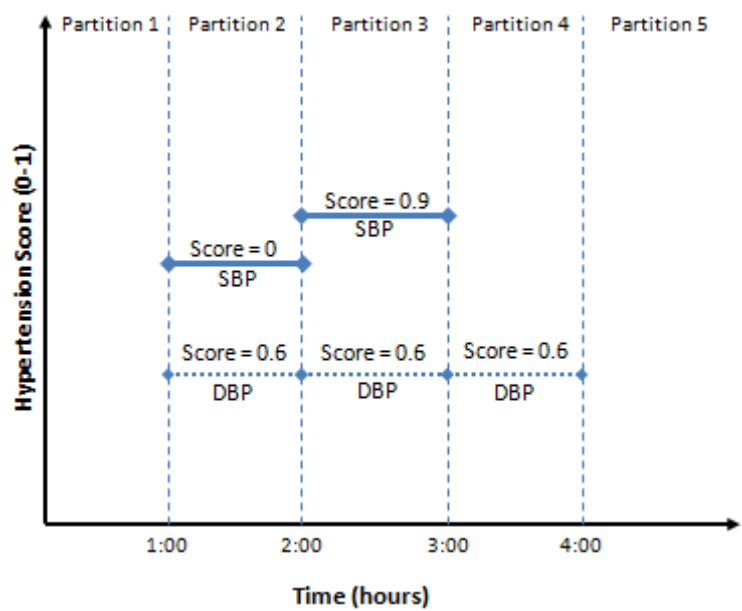

**Figure 9.** Evaluation of logical constraints by the Fuzzy Temporal Reasoner.

## Evaluation of Logic Operators

The last step of the reasoning process is the evaluation of the logic operators within compound logic expressions. For this, an additional fuzzy logic technique is used.

The operators **AND, OR,** and **NOT** of classic logic, exist in fuzzy logic with a different implementation. A fuzzy logic implementation of these logic operators, called *Zadeh operators,* uses the minimum function for the evaluation of the *AND* operator, and the maximum function for the evaluation of the *OR* operator, and uses the function 1-truth-value to evaluate the *NOT* operator.

### AND and OR Operators

In the Fuzzy Temporal Reasoner implementation we used the Zadeh operators for the evaluation of the **AND** and **OR** operators. The minimum function is used to evaluate the *AND* operator only in cases in which a membership score exists for all of the operands of the expression. The maximum function is used to evaluate the *OR* operator in cases in which the membership score is available for at least one of the operands of the expression.

Figure 10 illustrates evaluation of the logic operators on the demonstration example. The expression in the example is a compound expression using the *OR* operator (SBP>140 *OR* DBP>90); thus, for each of the partitions, the maximum function is applied on the membership scores that were calculated in the previous steps.

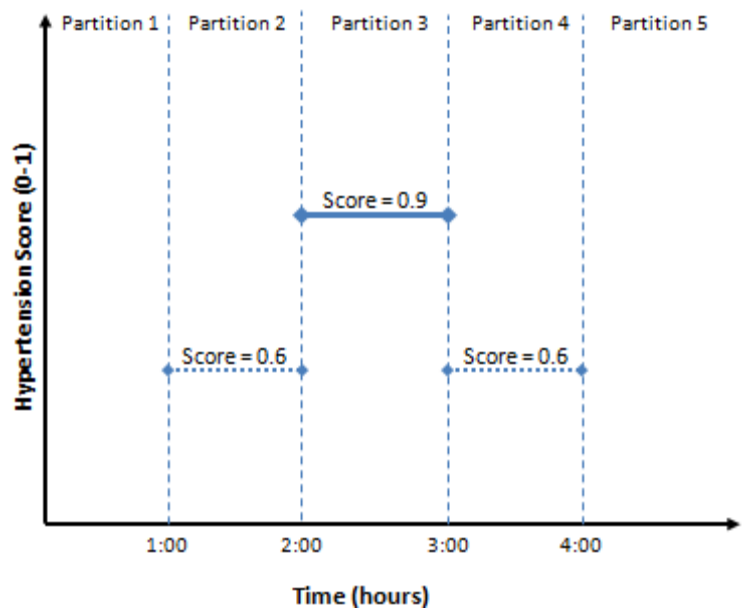

**Figure 10.** Evaluation of logic operators by the Fuzzy Temporal Reasoner*.*

The evaluation of the logic operators as described above enables the Fuzzy Temporal Reasoner to evaluate any complex compound expression. The knowledge specified in the KBTA schema allows the representation of complex compound logic expressions, represented in the form of *AND-OR* trees. For example, a definition of the Preeclampsia diagnosis is "high blood pressure with proteinuria in a pregnant woman after 20 weeks of gestation". Such definitions are specified in the knowledge as compound expressions, illustrated in Figure 11 using an *AND-OR* tree. The Fuzzy Temporal Reasoner uses a recursive implementation of the fuzzy logic evaluation function that supports evaluating any complex AND-OR tree.

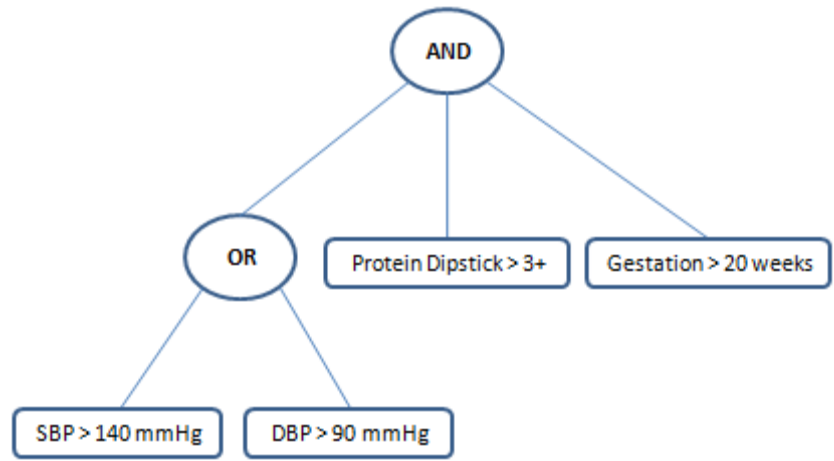

**Figure 11.** AND-OR tree representation of the Preeclampsia diagnosis concept.

**The NOT Operator**

For the **NOT** operator we have implemented an operation that inverts the relation by replacing the relation's operator with an opposite operator. For example, the expression *NOT*(*x* ≥ *y*) is evaluated as (*x* < *y*). The following example illustrates this new implementation. Consider the expression that defines a normal blood pressure condition as being *NOT*(*DBP* > 90 mmHg). The new implementation of the operator would set the (fuzzy) truth value to 1 whenever *DBP* ≤ *90.* When the values are between 90 to 100, the truth value would be set between 1 and 0, and whenever DBP ≥ 100, the truth value would be set to 0. Figure 12 demonstrates this concept.

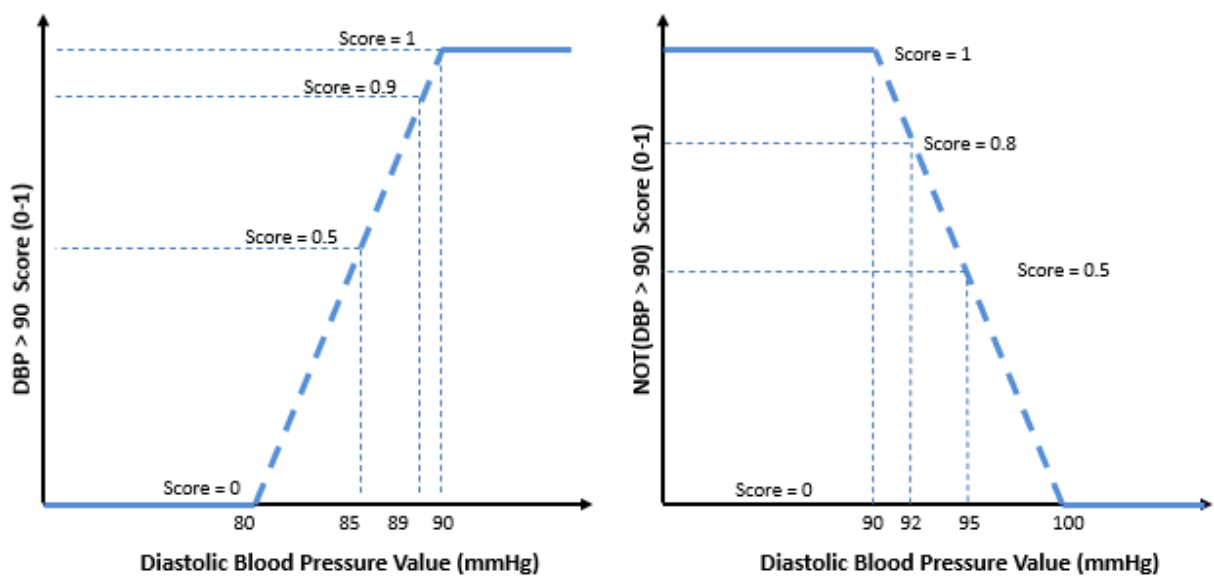

**Figure 12.** An example for fuzzy-evaluation of the NOT operator. The left graph displays the evaluation of the constraint DBP > 90mmHg, with deviation interval of 10mmHg. The right side displays the evaluation of the false-value of the same constraint, i.e., NOT(DBP > 90mmHg), assuming the same deviation interval.

To evaluate the fuzzy **false-value of compound operations**, we used an implementation of De Morgan's law. Evaluation of the expression *NOT(x OR y)*, is implemented as *(NOT(x) AND NOT(Y))*.

For example, to determine the fuzzy truth value of NOT having hypertension (which might be, for example, a stop condition for a therapy action), given the measurements of SBP=139 and DBP=92, with a deviation interval of 10mmHg, would be evaluated as:

*fuzzy-value(NOT((SBP ≥ 140) OR (DBP ≥ 90))) = fuzzy-value((SBP < 140) AND (DBP < 90)) = min(fuzzy-value(139), fuzzy-value(92)) = min(1,0.8) = 0.8*

Note that, assuming that the deviation interval for DBP ≥ 90mmHg is 10mmHg, the membership score of DBP=92 within the context of the fuzzy constraint DBP < 90, is 0.8 (see Figure 12).

## 2.3. The *DiscovErr* System

A concise summary of the overall architecture of the *DiscovErr* system, which fully implements the BiKBAC compliance-assessment methodology, is presented in Figure 13 (further technical details can be found elsewhere [Hatsek, 2014]). The *DiscovErr* system includes three main modules that are integrated to support the guideline compliance analysis task; the Knowledge Framework, including the knowledge library and knowledge specification tool, the Patient Data Access module, which retrieves the data from the electronic medical record, and the Analysis Framework, which performs the compliance analysis and provides a graphical interface for the users to view the compliance analysis results.

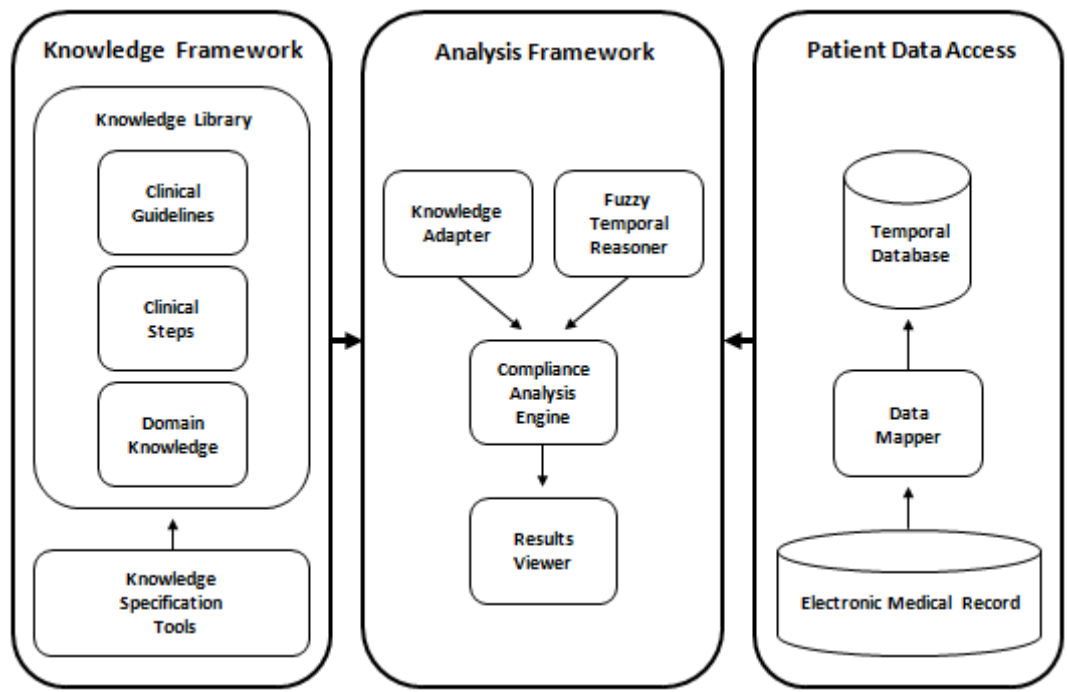

**Figure 13.** The overall architecture of the *DiscovErr* system

The Knowledge Framework includes a library that facilitates a knowledge model that integrates three types of formal knowledge: procedural knowledge of clinical guidelines (represented using the Asbru language), declarative medical domains knowledge (using the KBTA method), and knowledge about clinical steps with references to controlled medical vocabularies, such as the WHO's *Anatomical Therapeutic Chemical Classification System* (ATC) for drug related steps, and *Logical Observation*

*Identifiers Names and Codes* (LOINC) for laboratory test orders and results. The Knowledge Framework includes a graphical tool for knowledge specification, which is used by expert physicians and knowledge engineers to create and maintain the formal knowledge and store it in the knowledge library.

The Patient Data Access module includes the Data Mapper module for importing data from the electronic medical record and for storing it in a time-oriented (technically, historical) database according to the internal schema of the system. The Data Mapper performs three main tasks during the data import process: (1) map the concept identifiers used in the external electronic medical record to the concept identifiers used in the formal knowledge; (2) convert the data when required due to different units of measurement; and (3) store the time-oriented patient data in a time-oriented database, in which each raw data item (e.g., Hemoglobin level) is mapped to a corresponding concept in the formal knowledge, and includes the relevant valid-start and valid-stop time stamps. The data in the time-oriented database consists of two types of parameters; primitive parameters that represent information about the state of the patient (e.g., lab results, physical examinations) and event parameters that represent the medical treatment (e.g., medication orders, procedures).

The Analysis Framework is responsible for the computational task of compliance analysis. The core of the Analysis Framework is the Compliance Analysis Engine, which applies the computational algorithms to analyze the patient data regarding the compliance to the clinical guidelines. The analysis engine accesses the knowledge through the Knowledge Adapter that provides sophisticated methods to query the knowledge library. The analysis engine uses the Fuzzy Temporal Reasoner to extract high level temporal abstractions regarding the patient's state. In addition, the Analysis Framework includes the Results Viewer, a graphical interface that allows users to view and explore the compliance analysis results.

## 3. Evaluation of the *BIKBAC* Methodology and of the *DiscovErr* system

The *DiscovErr* system has been evaluated rigorously in a study that is outside of the scope of the current paper [Hatsek, 2014]. It was applied to 1584 records of 10 type 2 diabetes patients, and its comments were carefully compared to those of three medical experts who were all highly experienced in the management of diabetes patients (two of which are specifically diabetes experts).

The results of the evaluation had demonstrated that the completeness of the *DiscovErr* system was 91% when the gold standard was comments made by at least two of the three experts, and 98% when compared to comments made by all three experts. The correctness of the system was 91% when compared to comments judged as correct by one diabetes expert and at least as partially correct by the other. Both diabetes experts judged 89% of the comments made by the *DiscovErr* system as Important (versus the other option, Less Important); 8% were judged as Important by one of the two experts, and 3% were judged as Less Important by both experts.

Overall, if it were considered as an additional (fourth) expert, the *DiscovErr* system would have ranked first in completeness and second in correctness among the experts (when computing these measures also for the human experts, using a majority consensus of opinions as the gold standard).

# 4. Summary and Discussion

In the current study, we have introduced the *BiKBAC* methodology, a new approach for automated guideline-based quality assessment of the care process, which assesses the degree of compliance of the longitudinal treatment of a particular patient, with a given clinical guideline. The automated quality-assessment is performed through a highly detailed, automated retrospective analysis, which compares a full, formal representation of the guideline (using, in this case, the Asbru language), with the longitudinal electronic medical record of the patient and abstractions computed from it using the KBTA method. The comparison uses both a top-down and a bottom-up approach, which we have explained in detail. Partial matches of the data to the process and to the outcome objectives are resolved using fuzzy temporal reasoning. We also introduced the *DiscovErr* system, which implements the BiKBAC approach, presented its architecture, and briefly described its evaluation in the type 2 diabetes domain, with highly encouraging results.

## *Main Contributions of the Current Study*

Unlike previous approaches to the task of quality assessment, the methodology that we had presented exploits a full, formal representation of the clinical guideline, and quality measures that exploit full-blown temporal-abstraction patterns, based on all of the guideline's intermediate and overall process and outcome intentions, which represent a correct guideline-based process being carried out as planned, in a manner sensitive to the longitudinal, evolving contexts of each patient.

Furthermore, our methodology caters also for partial *(fuzzy temporal)* pattern matching of these temporally oriented objectives. Partial temporal-pattern matching of conditions and intentions is very important, since it is highly beneficial to detect deviations in compliance to a given guideline in a manner proportional to the level of the deviation, and not to simply use arbitrary cut-off values.

The critiquing is performed in a novel bidirectional fashion, using the Bidirectional Knowledge-Based Assessment of Compliance (BiKBAC) method. The new methodology exploits a formal representation of the evidence-based guideline, and uses both a top-down (driven by the guideline's process and outcome intentions, represented formally as Asbru temporally extended intentions) and a bottom-up (driven by the patient's multivariate longitudinal data and abstracted using the KBTA methodology) approach. As mentioned above, the need for avoiding overly crisp cut-off values is addressed in the BiKBAC methodology through the performance of partial matches of the data to the process and to the outcome objectives of the guideline, which are resolved using fuzzy temporal logic.

In the current study, we have also fully implemented the BIKBAC methodology within the *DiscovErr* system, whose architecture we had described in detail.

Finally, we have rigorously assessed (in a separate study) the *DiscovErr* system in the type 2 diabetes management domain, with the assistance of three domain experts, and producing highly encouraging results with respect to both completeness and correctness of the *DiscovErr* system in that domain.

The results of our research suggest the possibility of building and using systems such as the *DiscovErr* system, and the feasibility of representing formal procedural (e.g., Asbru) and declarative (e.g., KBTA) knowledge, and even the feasibility of exploiting fuzzy temporal logic, to build practical automated care-critiquing systems. Such systems might well be able to perform large-scale evidence-based quality assessment of large numbers of electronic medical records, given the necessary formal procedural and declarative knowledge representations.

Note that if sophisticated systems such as the *DiscovErr* system will be used for quality assessment, more sophisticated quality measures could be defined, to address other aspects of the correct application of the relevant medical knowledge; for example, the *temporal* aspects of the data (e.g., note that one might wish to examine the pattern formed by *all* of the LDL-C measurements, and not only the most recent one). Such a capability might make "rigging the system" by trying to adhere only to simplistic quality-assessment measures, much more difficult, and might encourage better adherence to the guideline.

Note also that the current study focused on the feasibility and validity of a *retrospective* automated critiquing process. The compliance analysis, however, can be performed in *real time*, at the point of care, by assessing the quality of the care provider's *decisions*, as opposed to *actions*, to *immediately* assist clinicians in increasing their compliance when deviations from the guidelines are being detected. Such a *critiquing mode*, "over the shoulder" style of guideline-based support aims to provide on-line decision support with minimal interaction with the clinician, thus enhancing the acceptance of decision-support systems in real clinical settings. Indeed, for that reason, earlier studies focused on critiquing, including systems such as *HyperCritic* [van der Lei and Musen 1990], *Trauma-TIQ* [Gertner 1997] *and AsthmaCritic* [Kuilboer et al. 2003].

## 5. Conclusions

We have presented a new, top-down and bottom-up quality-assessment methodology, the BIKBAC critiquing method, and implemented it as the *DiscovErr* system, for assessing the quality of evidence-based longitudinal care. The methodology is based on a formal representation of the evidence-based clinical guideline and its intentions, and on using fuzzy temporal logic to provide partial scores of adherence to the guideline's multiple types of formally represented conditions and intentions. The evaluation of the *DiscovErr* system in the type 2 diabetes management domain, given a set of type 2 diabetes patients managed for five years, comparing the system's comments to those of a panel of three clinical experts, was highly encouraging.

We conclude that the BIKBAC methodology is valid, and that systems such as the *DiscovErr* system are quite feasible to implement. We also conclude that the BIKBAC methodology can be effectively deployed, with respect to the necessary representation of procedural and declarative time-oriented clinical knowledge, and can be used to provide critique to complex, continuous, longitudinal medical care, such as type 2 diabetes management, at a level comparable to human domain experts.

## Acknowledgements

Dr. Hatsek's research was partially supported by the Israeli National Institute for Health Policy Research. We thank our three Diabetes experts, Drs. Hochberg, Daoud Naccache and Biderman for their assistance in evaluating the *DiscovErr* system.